\documentclass[10pt,a4paper, geometry, fullpage]{article}
\usepackage{booktabs}

\usepackage[utf8]{inputenc} % utf8 characters
\usepackage{amsmath} % math <3
\usepackage{amsfonts} % math fonts
\usepackage{amssymb} % math symbols
\usepackage{amsthm} % math theorems
\usepackage{graphicx} % to include images
\usepackage{epstopdf} % to compile eps. 
\usepackage{bbm} % enables \mathbbm{1}
\usepackage[lined,boxed,commentsnumbered]{algorithm2e}
\usepackage[toc]{appendix} % appendices
\usepackage[headsep=1cm,headheight=0cm]{geometry}
\usepackage{enumerate}
\usepackage{listings}
\usepackage{authblk}

	\usepackage[pdfborder={0 0 0},plainpages=false]{hyperref}

	\theoremstyle{plain}
	\theoremstyle{remark}
\title{Benchmarking the Hill-Valley Evolutionary Algorithm for the GECCO 2018 Competition on Niching Methods Multimodal Optimization}

\newcommand{\bx}{\mathbf{x}}

\newcommand{\by}{\mathbf{y}}

\newcommand{\cP}{\mathcal{P}}

\makeatletter
\def\Ddots{\mathinner{\mkern1mu\raise\p@
\vbox{\kern7\p@\hbox{.}}\mkern2mu
\raise4\p@\hbox{.}\mkern2mu\raise7\p@\hbox{.}\mkern1mu}}
\makeatother

\date{\today}
\author[1]{S.C. Maree}
\author[1]{T. Alderliesten}
\author[2]{D. Thierens}
\author[3]{P.A.N. Bosman}

\affil[1]{Amsterdam UMC, Amsterdam, The Netherlands}
\affil[2]{Utrecht University, Utrecht, The Netherlands}
\affil[3]{Centrum Wiskunde \& Informatica, Amsterdam, The Netherlands}

\begin{document}

\maketitle

\begin{abstract}
This report presents benchmarking results of the latest version of the Hill-Valley Evolutionary Algorithm (HillVallEA) on the CEC2013 niching benchmark suite. The benchmarking follows restrictions required by the GECCO 2018 competition on Niching methods for Multimodal Optimization. In particular, no problem dependent parameter tuning is performed. A number of adjustments have been made to original publication of HillVallEA that are discussed in this report.
\end{abstract}

\section{Introduction}
The Hill-Valley Evolutionary Algorithm (HillVallEA) \cite{Maree18} is a real-valued multi-modal evolutionary algorithm that automatically detects niches in the search space, based on a two-step approach. In the first step, the initial population is clustered into niches. The clustering is based on the Hill-Valley test. This test determines whether two solutions belong to the same niche (valley) by sampling and evaluating additional solutions along the line segment connecting the two solutions in the search space. In the second step,a population-based core search algorithm is initialized on each of the clusters. In this case, AMaLGaM-Univariate \cite{bosman13} is used, as it was the best performing core search algorithm \cite{Maree18}. We refer the interested reader to \cite{Maree18} for more details on the HillVallEA.

\section{Adaptations to the HillVallEA}

\subsection{Re-using solutions for parameter estimation after clustering}
Originally, the test solutions sampled in the Hill-Valley test were discarded, although these solutions may contain valuable information. When testing whether solution $\bx$ belongs to the cluster of another solution $\by$, we sample intermediate test solutions along the line segment between the two solutions in parameter space, starting at $\bx$. If $\bx$ is found to belong to the cluster of $\by$, we add $\bx$ and all test solutions to the cluster of $\by$. Otherwise, we store all test solutions, except for the one that violated the Hill-Valley test. Later, when $\bx$ is added to a (new) cluster, all stored test solutions are added to that cluster as well. 

\subsection{Termination criterion when re-exploring niches}
A termination criterion has been added that aims to terminate core search algorithms when a niche is being explored that was already explored previously. During a run of the HillVallEA, an elitist archive is kept with all distinct global optima that have been found so far. Every fifth generation of a core search algorithm, the best obtained solution is compared to the nearest solution in elitist archive using the Hill-Valley test (using five test solutions). When it is found that these two solutions belong to the same niche, the core search algorithm is terminated, as it is exploring a niche that was already explored previously. 

\subsection{Termination criterion when converging to a local minimum}
A second termination criterion that was added is aimed to detect whether a core search algorithm is converging to a local minimum. Let $b$ be the fitness value of the best solution in the elitist archive, which we use as a target value to compare future runs of core search algorithms against. 

AMaLGaM was shown to have exponential convergence to the global minimum $b$ on smooth unimodal functions such as the sphere function \cite{bosman13}. Let $a_{g}$ be the average fitness of the selection in generation $g$. Then,we define $\Delta_g := a_{g} - b$ as the distance to the global minimum. If $\Delta_g < 0$, the average fitness is better than the global minimum, and we do not terminate the core search algorithm. This could in practice happen as $b$ is not always known a priori, and needs to be approximated. For this benchmark, we assume $b$ is not known and we use the best elite in the elitist archive as approximation of $b$. 

Under the assumption of exponential convergence, $\Delta_{g+1}$ can be described in terms of $\Delta_g$ by 
\begin{equation}
\Delta_{g+1} = \Delta_g (1-r),
\end{equation}
where $r$ is the rate of convergence. We estimate $r$ by $r_n$ over the previous $n$ generations by,
\begin{equation}r \approx r_n = 1-\left(1-\frac{\Delta_{g-n} - \Delta_{g}}{\Delta _{g-n}}\right)^{1/n},
\end{equation}
with $n=5$ to reduce statistical noise. Note that when $r_n > 0$, the core search algorithm has improved average fitness. If $r_n \leq 0$, the algorithm is still in the exploratory phase. We therefore do not terminate it. To prevent premature termination, this termination criterion is only applied when $\Delta_g$ decreased consecutively in the most recent $n = 5$ generations consecutively. 

Finally, we estimate the \textit{time to optimum} ($tto$) in order to achieve $\Delta_{g+tto} = 10^{-12}$. Again, under the assumption of exponential convergence, $\Delta_{g+tto} = \Delta_g(1-r)^{tto}$. Rewriting this in terms of $tto$ gives,
\begin{equation}
tto = \frac{\log\left({10^ {-12}}/{\Delta_g }\right) }{\log\left(1-r\right)} \approx \frac{\log\left({10^ {-12}}/{\Delta_g }\right) }{\log\left(1-r_n\right)} = \frac{\log\left({10^ {-12}}/{\Delta_g }\right) }{\frac1n\log\left(1-\frac{\Delta_{g-n}-\Delta_g}{\Delta_{g-n}}\right)}
\end{equation}
A core search algorithm is terminated if $g + tto$ exceeds 50 times the maximum number of generations it took to find any elite in the elitist archive.

\subsection{Recalibration of the recursion scheme}
After adapting the termination criteria for the core search algorithms, a novel parameter setting of the population-size growing scheme was found to enhance performance. Specifically, by increasing the initial population size to $|\cP| = 2^6d$, where $d$ is the problem dimensionality, overall performance increased (while $|\cP| = 2^ 4d$ was proposed in the original HillVallEA). 

\section{Experiments}
We evaluate the performance of the adapted version of the HillVallEA on the CEC2013 niching benchmark suite \cite{CEC2013NichingCompetition}. The benchmark consists of 20 problems, as shown in Table~\ref{tab:2dbenchmarks}, to be solved within a predefined budget in terms of function evaluations. For each of the benchmark problems, the location of the optima and the corresponding fitness values are known. However, these are only used for measure performance, and are thus not used by the HillVallEA. 

Two performance measures are used in this work. The peak ratio (PR) measures the fraction of global optima detected, computed according to the guidelines of the niching benchmark suite. In contrast to the full benchmark suite, where a range of accuracies $\varepsilon$ is used, and the final peak ratio is the mean of these results, we only use the highest accuracy of $\varepsilon = 10^{-5}$. Due to the post-processing step in HillVallEA, all local optima are filtered out, making the results independent of the choice of $\varepsilon \geq 10^{-5}$.

The second measure that is used is the static $F_1$ measure. The static $F_1$ measure is the ratio of solutions that turn out to be distinct global optima out of the full set of reported solutions.

The GECCO 2018 competition on Niching Methods for Multimodal Optimization uses a third performance measure, which is the Dynamic $F_1$ measure. The dynamic $F_1$ measure is the area under the curve of the static $F_1$ measure over time (measured function evaluations). However, specific implementation details are unknown, thus this measure is not incorporated in this work. 

All benchmark functions are defined on a bounded domain. All experiments in this work are repeated 50 times, and resulting performance measures are averaged over all repetitions. No problem-specific parameter tuning has been performed.

\begin{table}
\begin{center}
\small
\begin{tabular}{lccccc|cc|cc}
\toprule
\multicolumn{6}{c|}{Benchmark details} & \multicolumn{2}{c|}{RS-CMSA} & \multicolumn{2}{c}{HillVallEA} \\
\# & Function name & $d$ & \#gopt & \#lopt & budget & PR & $F_1$ & PR & $F_1$ \\
\toprule
1 & Five-Uneven-Peak Trap & 1 & 2 & 3 & 50K& 1& 0.99& 1 & 1  \\
2 & Equal Maxima & 1 & 5 & 0 & 50K& 1&1& 1 & 1  \\
3 & Uneven Decreasing Maxima & 1 & 1 & 4 & 50K&1& 0.98 & 1 & 1 \\
4 & Himmelblau & 2 & 4 & 0 & 50K &1  & 1&1 & 1 \\
5 & Six-Hump Camel Back & 2 & 2 & 5 & 50K& 1&1& 1 & 1  \\
6 & Shubert & 2 & 18 & many & 200K & 0.999 &1&  1 & 1  \\
7 & Vincent & 2 & 36 & 0 & 200K & 0.997 &1& 1 & 1\\
8 & Shubert & 3 & 81 & many & 400K & 0.871 & 1&0.920 & 1 \\
9 & Vincent & 3 & 216 & 0 & 400K & 0.730 &1& 0.945 & 1 \\
10 & Modified Rastrigin & 2 & 12 & 0 & 200K & 1 &1& 1 & 1 \\
11 & Composition Function 1 & 2 & 6 & many &200K & 0.997 &1& 1 & 1\\
12 & Composition Function 2 & 2 & 8 &many & 200K & 0.948 &1& 1 & 1\\
13 & Composition Function 3 & 2 & 6 &many & 200K & 0.997 &1& 1 & 1 \\
14 & Composition Function 3 & 3 & 6 &many & 400K & 0.810 & 1&0.917 & 1 \\
15 & Composition Function 4 & 3 & 8 &many & 400K & 0.748 &1& 0.750 & 1 \\
16 & Composition Function 3 & 5 & 6 &many & 400K & 0.667 &1& 0.687 & 1\\
17 & Composition Function 4 & 5 & 8 &many & 400K & 0.703 &1& 0.750 & 1\\
18 & Composition Function 3 & 10 & 6 & many &400K & 0.667 &1& 0.667 & 1 \\
19 & Composition Function 4 & 10 & 8 &many & 400K & 0.503 & 0.996 & 0.585 & 1 \\
20 & Composition Function 4 & 20 & 8 &many & 400K & 0.483 &1& 0.482 &1\\
\bottomrule 
\multicolumn{6}{r|}{average:} & 0.856 &  0.998 & 0.885 & 1 \\
\end{tabular}
\caption{Niching benchmark suite from the CEC2013 special session on multi-modal optimization \cite{CEC2013NichingCompetition}. For each problem the function name, problem dimensionality $d$, number of global optima $\#gopt$, number of local optima $\#lopt$, and budget in terms of function evaluations are given. Peak ratio (PR) is fraction of obtained global optima. Static $F_1$ is the fraction of distinct global optima in the resulting solution set.}
\label{tab:2dbenchmarks}
\end{center}
\end{table}

\section{Results and discussion}
Table~\ref{tab:2dbenchmarks} shows the obtained PR and static $F_1$ for each test problem, averaged over 50 independent runs. Results of RS-CMSA \cite{ahrari17}, winner of the GECCO'17 niching competition, are shown for comparison. HillVallEA successfully filters out all duplicate optima, as the static $F_1$ measure is 1 for all instances. Furthermore, HillVallEA outperforms RS-CMSA on average in peak ratio. 

Source code (C++) of the HillVallEA to reproduce the experiments in this report is available at \url{https://github.com/SCMaree/HillVallEA}.

{\small
\bibliographystyle{unsrt}
\bibliography{Maree_Gecco18b}
}
\end{document}